\documentclass[11pt,a4paper]{article}
\usepackage{acl2015}
\usepackage{times}
\usepackage{url}
\usepackage{latexsym}
\usepackage{amsmath}
\usepackage{enumitem}
\usepackage{lipsum}
\usepackage{graphicx}
\usepackage{subcaption}
\usepackage{caption}
\usepackage{tabularx}
\usepackage{color,soul}
\usepackage{xcolor}
\usepackage{multirow}
\usepackage{booktabs}

\makeatletter
\newcommand{\@BIBLABEL}{\@emptybiblabel}
\newcommand{\@emptybiblabel}[1]{}
\makeatother
\PassOptionsToPackage{hidelinks}{hyperref}
\usepackage{hyperref}

\graphicspath{ {fig/} }

\newcommand*{\V}[1]{\mathbf{#1}}%
\definecolor{orange-light}{HTML}{FFC7E9}
\definecolor{green-light}{HTML}{EAFFD1}
\definecolor{blue-light}{HTML}{C9EAFF}
\newcommand{\hlco}[2][orange-light]{ {\sethlcolor{#1} \hl{#2}} }
\newcommand{\hlg}[2][green-light]{ {\sethlcolor{#1} \hl{#2}} }
\newcommand{\hlcb}[2][blue-light]{ {\sethlcolor{#1} \hl{#2}} }

\title{Scientific Article Summarization Using Citation-Context \\ and Article's Discourse Structure}

\author{Arman Cohan \\
  Information Retrieval Lab \\
  Computer Science Department \\
  Georgetown University \\
  {\tt arman@ir.cs.georgetown.edu} \\\And
  Nazli Goharian \\
  Information Retrieval Lab \\
  Computer Science Department \\
  Georgetown University \\
  {\tt nazli@ir.cs.georgetown.edu} \\}

\date{}

\begin{document}
\maketitle
\begin{abstract}
  We propose a summarization approach for scientific articles which takes advantage of citation-context and the document discourse model. While citations have been previously used in generating scientific summaries, they lack the related context from the referenced article and therefore do not accurately reflect the article's content.
   Our method overcomes the problem of inconsistency between the citation summary and the article's content by providing context for each citation. We also leverage the inherent scientific article's discourse for producing better summaries. We show that our proposed method effectively improves over existing summarization approaches (greater than 30\% improvement over the best performing baseline) in terms of \textsc{Rouge} scores on TAC2014 scientific summarization dataset. While the dataset we use for evaluation is in the biomedical domain, most of our approaches are general and therefore adaptable to other domains.
\end{abstract}


\section{Introduction}
Due to the expanding rate at which articles are being published in each scientific field, it has become difficult for researchers to keep up with the developments in their respective fields.
Scientific summarization aims to facilitate this problem by providing readers with concise and informative representation of contributions or findings of an article. Scientific summarization is different than general summarization in three main aspects \cite{Teufel:2002}. First, the length of scientific papers are usually much longer than general articles (e.g newswire). Second, in scientific summarization, the goal is typically to provide a technical summary of the paper which includes important findings, contributions or impacts of a paper to the community. Finally, scientific papers follow a natural discourse. A common organization for scientific paper is the one in which the problem is first introduced and is followed by the description of hypotheses, methods, experiments, findings and finally results and implications. Scientific summarization was recently further motivated by TAC2014 biomedical summarization track\footnotemark[1] in which they planned to investigate this problem in the domain of biomedical science.
\footnotetext[1]{Text Analysis Conference - \url{http://www.nist.gov/tac/2014}}

There are currently two types of approaches towards scientific summarization. First is the articles' abstracts. While abstracts provide a general overview of the paper, they cannot be considered as an accurate scientific summary by themselves. That is due to the fact that not all the contributions and impacts of the paper are included in the abstract \cite{elkiss2008blind}. In addition, the stated contributions are those that the authors deem important while they might be less important to the scientific community. Moreover, contributions are stated in a general and less focused fashion.
These problems motivated the other form of scientific summaries, i.e., citation based summaries. Citation based summary is a summary which is formed by utilizing a set of citations to a referenced article \cite{qazvinian2008scientific,qazvinian2013generating}. This set of citations has been previously indicated as a good representation of important findings and contributions of the article. Contributions stated in the citations are usually more focused than the abstract and contain additional information that is not in the abstract \cite{elkiss2008blind}.

However, citations may not accurately represent the content of the referenced article as they are biased towards the viewpoint of the citing authors. Moreover, citations may address a contribution or a finding regarding the referenced article without referring to the assumptions and data under which it was obtained.

The problem of inconsistency between the degree of certainty of expressing findings between the citing article and referenced article has been also reported \cite{deWaard2012epistemic}. Therefore, citations by themselves lack the related ``context'' from the original article. We call the textual spans in the reference articles that reflect the citation, the citation-context. Figure \ref{fig:context} shows an example of the citation-context in the reference article (green color) for a citation in the citing article (blue color).

We propose an approach to overcome the aforementioned shortcomings of existing scientific summaries. Specifically, we extract citation-context in the reference article for each citation. Then, by using the discourse facets of the citations as well as community structure of the citation-contexts, we extract candidate sentences for the summary. The final summary is formed by maximizing both novelty and informativeness of the sentences in the summary. We evaluate and compare our methods against several well-known summarization methods. Evaluation results on the TAC2014 dataset show that our proposed methods can effectively improve over the well-known existing summarization approaches. That is, we obtained greater than 30\% improvement over the highest performing baseline in terms of mean \textsc{Rouge} scores.

\begin{figure}[t]
\scriptsize
Citing article: \\
\fbox{\begin{minipage}{0.97\linewidth}
 ...
 \hlcb{The general impression that has emerged is that transformation of human cells by Ras requires the inactivation of both the pRb and p53 pathways, typically achieved by introducing DNA tumor virus oncoproteins such as SV40 large tumor antigen (T-Ag) or human papillomavirus E6 and E7 proteins} (\hlco{Serrano et al., 1997}).\\
To address this question, we have been investigating the ...

\end{minipage}} \\
Reference article (Serrano et al., 1997): \\
\fbox{\begin{minipage}{0.97\linewidth}
 ... continued to incorporate BrdU and proliferate following introduction of H-ras V12. \hlg{ In agreement with previous reports ( 66 and 60), both p53−/− and p16−/− MEFs expressing H-ras V12 displayed features of oncogenic transformation (e.g., refractile morphology, loss of contact inhibition), which were apparent almost immediately after H-ras V12 was transduced (data not shown). These results indicate that p53 and p16 are essential for ras-induced arrest in MEFs, and that inactivation of either p53 or p16 alone is sufficient to circumvent arrest.}
In REF52 and IMR90 fibroblasts, a different approach was ...
\end{minipage}}
\caption{\footnotesize The blue highlighted span in the citing article (top) shows the citation text, followed by the citation marker (pink span). For this citation, the citation-context is the green highlighted span in the reference article (bottom). The text spans outside the scope of the citation text and citation-context are not highlighted. }
\label{fig:context}
\end{figure}


\section{Related work}
\label{sec:related}

Document summarization is a relatively well studied area and various types of approaches for document summarization have been proposed in the past twenty years.

Latent Semantic Analysis (LSA) has been used in text summarization first by \newcite{gong2001lsa}.  Other variations of LSA based summarization approaches have later been introduced \cite{steinberger2004lsa,steinberger2005improving,Lee200920,ozsoy2010text}. Summarization approaches based on topic modeling and Bayesian models have also been explored \cite{vanderwende2007,haghighi2009,celikyilmaz2010hybrid,Ritter:2010:unsupervised,celikyilmaz2011discovery,ma2013,li2014novel}. In these approaches, the content/topic distribution in the final summary is estimated using a graphical probabilistic model. Some approaches have viewed summarization as an optimization task solved by linear programming \cite{Clarke:2008,berg2011jointly,D12-1022}. Many works have viewed the summarization problem as a supervised classification problem in which several features are used to predict the inclusion of document sentences in the summary. Variations of supervised models have been utilized for summary generation, such as: maximum entropy \cite{osborne2002using}, HMM \cite{conroy2011classy}, CRF \cite{galley2006skip,shen2007document,Chali:2012:QMS:2139643.2139649}, SVM \cite{Xie2010495}, logistic regression \cite{louis2010discourse} and reinforcement learning \cite{rioux2014}. Problems with supervised models in context of summarization include the need for large amount of annotated data and domain dependency.

Graph based models have shown promising results for text summarization. In these approaches, the goal is to find the most central sentences in the document by constructing a graph in which nodes are sentences and edges are similarity between these sentences. Examples of these techniques include LexRank \cite{erkan2004}, TextRank \cite{mihalcea2004}, and the work by \newcite{Paul:2010}. Maximizing the novelty and preventing the redundancy in a summary is addressed by greedy selection of content summarization \cite{carbonell1998use,guo2010probabilistic,lin2010putting}. Rhetorical structure of the documents have also been investigated for automatic summarization. In this line of work, dependency and discourse parsing based on Rhetorical Structure Theory (RST) \cite{mann1988rhetorical} is used for analyzing the structure of the documents \cite{hirao2013single,kikuchi2014single,yoshida2014dependency}. Summarization based on rhetorical structure is better suited for shorter documents and is highly dependent on the quality of the discourse parser that is used. Training the discourse parser requires large amount of training data in the RST framework.

Scientific article summarization was first studied by \newcite{Teufel:2002} in which they trained a supervised Naive Bayes classifier to select informative content for the summary. Later \newcite{elkiss2008blind} argue the benefits of citations to scientific work analysis. \newcite{cohan2015} use a search oriented approach for finding relevant parts of the reference paper to citations. \newcite{qazvinian2008scientific,qazvinian2013generating} use citations to an article to construct its summary. More specifically, they perform hierarchical agglomerative clustering on citations to maximize purity and select most central sentences from each cluster for the final summary. Our work is closest to \newcite{qazvinian2008scientific} with the difference that they only make use of citations.
While citations are useful for summarization, relying solely on them might not accurately capture the original context of the referenced paper. That is, the generated summary lacks the appropriate evidence to reflect the content of the original paper, such as circumstances, data and assumptions under which certain findings were obtained. We address this shortcoming by leveraging the citation-context and the inherent discourse model in the scientific articles.




\section{The summarization approach}
\label{sec:method}

Our scientific summary generation algorithm is composed of four steps: (1) Extracting the citation-context, (2) Grouping citation-contexts, (3) Ranking the sentences within each group and (4) Selecting the sentences for final summary. We assume that the citation text (the text span in the citing article that references another article) in each citing article is already known. We describe each step in the following sub-sections. Our proposed method generates a summary of an article with the premise that the article has a number of citations to it. We call the article that is being referenced the ``reference article''.
We shall note that we tokenized the articles' text to sentences by using the punkt unsupervised sentence boundary detection algorithm \cite{kiss2006unsupervised}. We modified the original sentence boundary detection algorithm to also account for biomedical abbreviations. For the rest of the paper, ``sentence'' refers to units that are output of the sentence boundary detection algorithm, whereas ``text span'' or in short ``span'' can consist of multiple sentences.

\subsection{Extracting the citation-context}
\label{subsec:method1}
As described in section \ref{sec:related}, one problem with existing citation based summarization approaches is that they lack the context of the referenced paper. Therefore, our goal is to leverage citation-context in the reference article to correctly reflect the reference paper. To find citation-contexts, we consider each citation as an n-gram vector and use vector space model for locating the relevant text spans in the reference article. More specifically, given a citation $c$, we return the ranked list of text spans $r_1,r_2,...,r_n$ which have the highest similarity to $c$. We call the retrieved text spans reference spans. These reference spans are essentially forming the context for each citation. The similarity function is the cosine similarity between the pivoted normalized vectors. We evaluated four different approaches for forming the citation vector.
\begin{enumerate}[topsep=0pt,itemsep=-1ex,partopsep=1ex,parsep=1ex,leftmargin=0ex,itemindent=4ex]
\item All terms in citation except for stopwords, numeric values and citation markers i.e., name of authors or numbered citations. In figure \ref{fig:context} an example of citation marker is shown.
\item Terms with high inverted document frequency (idf). Idf values of terms have shown to be a good estimate of term informativeness.
\item Concepts that are represented through noun phrases in the citation, for example in the following: `` ... typically achieved by introducing DNA tumor virus oncoproteins such a ... '' which is part of a citation, the phrase ``DNA tumor virus oncoproteins'' is a noun phrase.
\item Biomedical concepts and noun phrases expanded by related biomedical concepts: This formation is specific to the biomedical domain. It selects biomedical concepts and noun phrases in the citation and uses related biomedical terminology to expand the citation vector. We used Metamap\footnotemark[1] for extracting biomedical concepts from the citation text (which is a tool for mapping free form text to \textsc{umls}\footnotemark[2] concepts). For expanding the citation vector using the related biomedical terminology, we used \textsc{SNOMED CT}\footnotemark[3] ontology by which we added synonyms of the concepts in the citation text to the citation vector.
\end{enumerate}

\footnotetext[1]{\url{http://metamap.nlm.nih.gov/}}
\footnotetext[2]{Unified Medical Language System - a compendium of controlled vocabularies in the biomedical sciences, \url{http://www.nlm.nih.gov/research/umls} }
\footnotetext[3]{\url{http://www.nlm.nih.gov/research/umls/Snomed/snomed_main.html}}

\subsection{Grouping the citation-contexts}
\label{sec:group}
After identifying the context for each citation, we use them to form the summary. To capture various important aspects of the reference article, we form groups of citation-contexts that are about the same topic. We use the following two approaches for forming these groups:

\begin{description}[topsep=0pt,itemsep=1ex,partopsep=1ex,parsep=1ex,leftmargin=0ex]
\item[Community detection - ]
We want to find diverse key aspects of the reference article. We form the graph of extracted reference spans in which nodes are sentences and edges are similarity between sentences. As for the similarity function, we use cosine similarity between tf-idf vectors of the sentences. Similar to \cite{qazvinian2008scientific}, we want to find subgraphs or communities whose intra-connectivity is high but inter-connectivity is low. Such quality is captured by the modularity measure of the graph \cite{newman2006modularity,newman2012communities}. Graph modularity quantifies the denseness of the subgraphs in comparison with denseness of the graph of randomly distributed edges and is defined as follows:
\begin{equation*}
Q=\frac{1}{2m}\sum_{vw}\big[A_{vw}-\frac{k_v\times k_w}{2m}\big] \delta(c_v,c_w)
\end{equation*}
Where $A_{vw}$ is the weigh of the edge $(v,w)$; $k_v$ is the degree of the vertex $v$; $c_v$ is the community of vertex $v$; $\delta$ is the Kronecker's delta function and $m=\sum_{vw}A_{vw}$ is the normalization factor.

While the general problem of precise partitioning of the graph into highly dense communities that optimizes the modularity is computationally prohibitive \cite{brandes2008modularity}, many heuristic algorithms have been proposed with reasonable results. To extract communities from the graph of reference spans, we use the algorithm proposed by \newcite{blondel2008fast} which is a simple yet accurate and efficient community detection algorithm. Specifically, communities are built in a hierarchical fashion. At first, each node belongs to a separate community. Then nodes are assigned to new communities if there is a positive gain in modularity. This process is applied iteratively until no further improvement in modularity is possible.

\item[Discourse model - ] A natural discourse model is followed in each scientific article. In this method, instead of finding communities to capture different important aspects of the paper, we try to select reference spans based on the discourse model of the paper. The discourse model is according to the following facets: ``hypothesis'', ``method'', ``results'', ``implication'', ``discussion'' and ``data-set-used''.
The goal is to ideally include reference spans from each of these discourse facets of the article in the summary to correctly capture all aspects of the article.
We use a one-vs-rest SVM supervised model with linear kernel to classify the reference spans to their respective discourse facets. Training was done on both the citation and reference spans since empirical evaluation showed marginal improvements upon including the reference spans in addition to the citation itself. We use unigram and verb features with tf-idf weighting to train the classifier.

\subsection{Ranking model}
To identify the most representative sentences of each group, we require a measure of importance of sentences. We consider the sentences in a group as a graph and rank nodes based on their importance. An important node is a node that has many connections with other nodes. There are various ways of measuring centrality of nodes such as nodes’ degree, betweenness, closeness and eigenvectors. Here, we opt for eigenvectors and we find the most central sentences in each group by using the ``power method'' \cite{erkan2004} which iteratively updates the eigenvector until convergence.

\end{description}
\subsection{Selecting the sentences for final summary}
After scoring and ranking the sentences in each group which were identified either by discourse model or by community detection algorithm, we employ two strategies for generating the summary within the summary length threshold. 
\begin{itemize}[topsep=0pt,itemsep=-1ex,partopsep=1ex,parsep=1ex,leftmargin=1ex]
\item{\textit{Iterative}:} We select top sentences iteratively from each group until we reach the summary length threshold. That is, we first pick the top sentence from all groups and if the threshold is not met, we select the second sentence and so forth. In the discourse based method, the following ordering for selecting sentences from groups is used: ``hypothesis'', ``method'',``results'', ``implication'' and ``discussion''. In the community detection method, no pre-determined order is specified.

\item{\textit{Novelty:}} We employ a greedy strategy similar to MMR \cite{carbonell1998use} in which sentences from each group are selected based on the following scoring formula:
\begin{align*}
\text{score(S)} \stackrel{\text{def}}{=} & \lambda Sim_1\text{(S, D)} \\
& - (1-\lambda)Sim_2\text{(S,Summary)}
\end{align*}
Where, for each sentence $S$, the score is a linear interpolation of similarity of sentence with all other sentences ($Sim_1$) and the similarity of sentence with the sentences already in the summary ($Sim_2$) and $\lambda$ is a constant. We empirically set $\lambda=0.7$ and also selected top 3 central sentences from each group as the candidates for the final summary.
\end{itemize}


\section{Experimental setup}
\label{sec:exper}

\subsection{Data}
We used the TAC2014 biomedical summarization dataset for evaluation of our proposed method. The TAC2014 benchmark contains 20 topics each of which consists of one reference article and several articles that have citations to each reference article (the statistics of the dataset is shown in Table \ref{tab:data1}). All articles are biomedical papers published by Elsevier. For each topic, 4 experts in biomedical domain have written a scientific summary of length not exceeding 250 words for the reference article. The data also contains annotated citation texts as well as the discourse facets. The latter were used to build the supervised discourse model. The distribution of discourse facets is shown in Table \ref{tab:data2}.

\begin{table}[t]
\caption{\small Dataset statistics}
\label{tab:data1}
\footnotesize
\vspace{-1em}
\resizebox{\linewidth}{!}{%
\begin{tabular}{@{}lrr@{}}
  \toprule
Statistic                              & mean           & std        \\ \midrule
\# of topics (reference articles)              & 20             & 0          \\
\# of Gold summaries for each topic & 4              & 0          \\
\# of citing articles in each topic    & 15.65          &    2.70  \\
\# of citations to the reference article in  \\ \hspace{1.3em} each citing article    &  \multirow{2}{*}[15pt]{1.57}          &    \multirow{2}{*}[15pt]{1.17}  \\
Length of summaries (words)   & 235.64         & 31.24         \\
Length of articles (words)   & 9759.86    & 2199.48 \\ \bottomrule
\end{tabular}
}
\vspace{-0.5em}
\end{table}

\begin{table}[t]
\footnotesize
\centering
\caption{\small Distribution of annotated discourse facets}
\label{tab:data2}
\vspace{-0.5em}
\vspace{-1pt}
\setlength{\tabcolsep}{25pt}
\begin{tabular}{@{}lr@{}}
  \toprule
Discourse facet     & count         \\ \midrule
Hypothesis      & 21            \\
Method              & 155           \\
Results       &  490      \\
Implication       & 140           \\
Discussion        & 446           \\ \bottomrule
\end{tabular}
\end{table}


\subsection{Baselines}
\label{subsec:exper-lr}
We compared existing well-known and widely-used approaches discussed in section \ref{sec:related} with our approach and evaluated their effectiveness for scientific summarization. The first three approaches use the scientific article's text and the last approach uses the citations to the article for generating the summary. 
\begin{itemize}[topsep=0pt,itemsep=-1ex,partopsep=1ex,parsep=1ex,leftmargin=0ex,itemindent=1ex]
\item{\textit{LSA \cite{steinberger2004lsa}} - } The LSA summarization method  is based on singular value decomposition. In this method, a term document index $\V{A}$ is created in which the values correspond to the tf-idf values of terms in the document. Then, Singular Value Decomposition, a dimension reduction approach, is applied to $A$. This will yield a singular value matrix $\V{\Sigma}$ and a singular vector matrix $\V{V}^T$. The top singular vectors are selected from $\V{V}^T$ iteratively until length of the summary reaches a predefined threshold.
\item{\textit{LexRank \cite{erkan2004}} - } LexRank uses a measure called centrality to find the most representative sentences in given sets of sentences. It finds the most central sentences by updating the score of each sentence using an algorithm based on PageRank random walk ranking model \cite{page1999pagerank}. More specifically, the centrality score of each sentence is represented by a centrality matrix $\V{p}$ which is updated iteratively through the following equation using a method called ``power method'':
\begin{equation*}
\V{p}=\V{A}^T\V{p}
\end{equation*}
Where matrix $\V{A}$ is based on the similarity matrix $\V{B}$ of the sentences:
$$\V{A}=[d\V{U}+(1-d)\V{B}]$$
In which $\V{U}$ is a square matrix with values $1/N$ and $d$ is a parameter called the damping factor. We set $d$ to 0.1 which is the default suggested value.

\item{\textit{MMR \cite{carbonell1998use}} - } In Maximal Marginal Relevance (MMR), sentences are greedily ranked according to a score based on their relevance to the document and the amount of redundant information they carry. It scores sentences based on the maximization of the linear interpolation of the relevance to the document and diversity: \
\begin{align*}
\text{MMR(S,D)} \stackrel{\text{def}}{=} & \lambda Sim_1\text{(S, D)} \\
& - (1-\lambda)Sim_2\text{(S,Summary)}
\end{align*}
Where $S$ is the sentence being evaluated, $D$ is the document being summarized, $Sim_1$ and $Sim_2$ are similarity function, $Summary$ is the summary formed by the previously selected sentences and $\lambda$ is a parameter. We used cosine similarity as similarity functions and we set $\lambda$ to 0.3, 0.5 and 0.7 for observing the effect of informativeness vs. novelty.

\item{\textit{Citation summary \cite{qazvinian2008scientific}}- } In this approach, a network of citations is built and citations are clustered to maximum purity \cite{zhao2001criterion} and mutual information. These clusters are then used to generate the final summary by selecting the top central sentences from each cluster in a round-robin fashion. Our approach is similar to this work in that they also use centrality scores on citation network clusters. Since they only focus on citations, comparison of our approach with this work gives a better insight into how beneficial our use of citation-context and article's discourse model can be in generating scientific summaries.

\end{itemize}


\section{Results and discussions}

\subsection{Evaluation metrics}
We use the \textsc{Rouge} evaluation metrics which has shown consistent correlation with manually evaluated summarization scores \cite{lin2004rouge}. More specifically, we use \mbox{\textsc{Rouge-L}}, \mbox{\textsc{Rouge-1}} and \mbox{\textsc{Rouge-2}} to evaluate and compare the quality of the summaries generated by our system. While \mbox{\textsc{Rouge-N}} focuses on n-gram overlaps, \textsc{Rouge-L} uses the longest common subsequence to measure the quality of the summary. \textsc{Rouge-N} where $N$ is the n-gram order, is defined as follows:
\begin{align*}
\textsc{Rouge-N}=\frac{\sum\limits_{S\in \text{\{Gold summaries\}}}\sum\limits_{W \in S}f_{match}(W)}{\sum\limits_{S\in \text{\{Gold summaries\}}}\sum\limits_{W\in S}f(W)}
\end{align*}
Where $W$ is the n-gram, $f(.)$ is the count function, $f_{match}(.)$ is the maximum number of n-grams co-occurring in the generated summary and in a set of gold summaries. For a candidate summary $C$ with $n$ words and a gold summary $S$ with $u$ sentences, \textsc{Rouge-L} is defined as follows:
\vspace{-2pt}
\begin{align*}
\textsc{Rouge-L$_{rec}$}=\frac{\sum\limits_{i=1}^{u}LCS_\cup(r_i,C)}{\sum_{i=1}^{u}\vert r_i\vert}
\end{align*}
\begin{align*}
\textsc{Rouge-L$_{prec}$}=\frac{\sum\limits_{i=1}^{u}LCS_\cup(r_i,C)}{n}
\end{align*}

Where $LCS_\cup(.,.)$ is the Longest common subsequence (LCS) score of the union of LCS between gold sentence $r_i$ and the candidate summary $C$. \textsc{Rouge-L} f score is the harmonic mean between precision and recall.

\subsection{Comparison between summarizers}

\begin{figure*}
\centering
\begin{minipage}[t]{\textwidth}
\hspace{1em}
\includegraphics[width=0.5\textwidth, height=0.295\textheight,keepaspectratio]{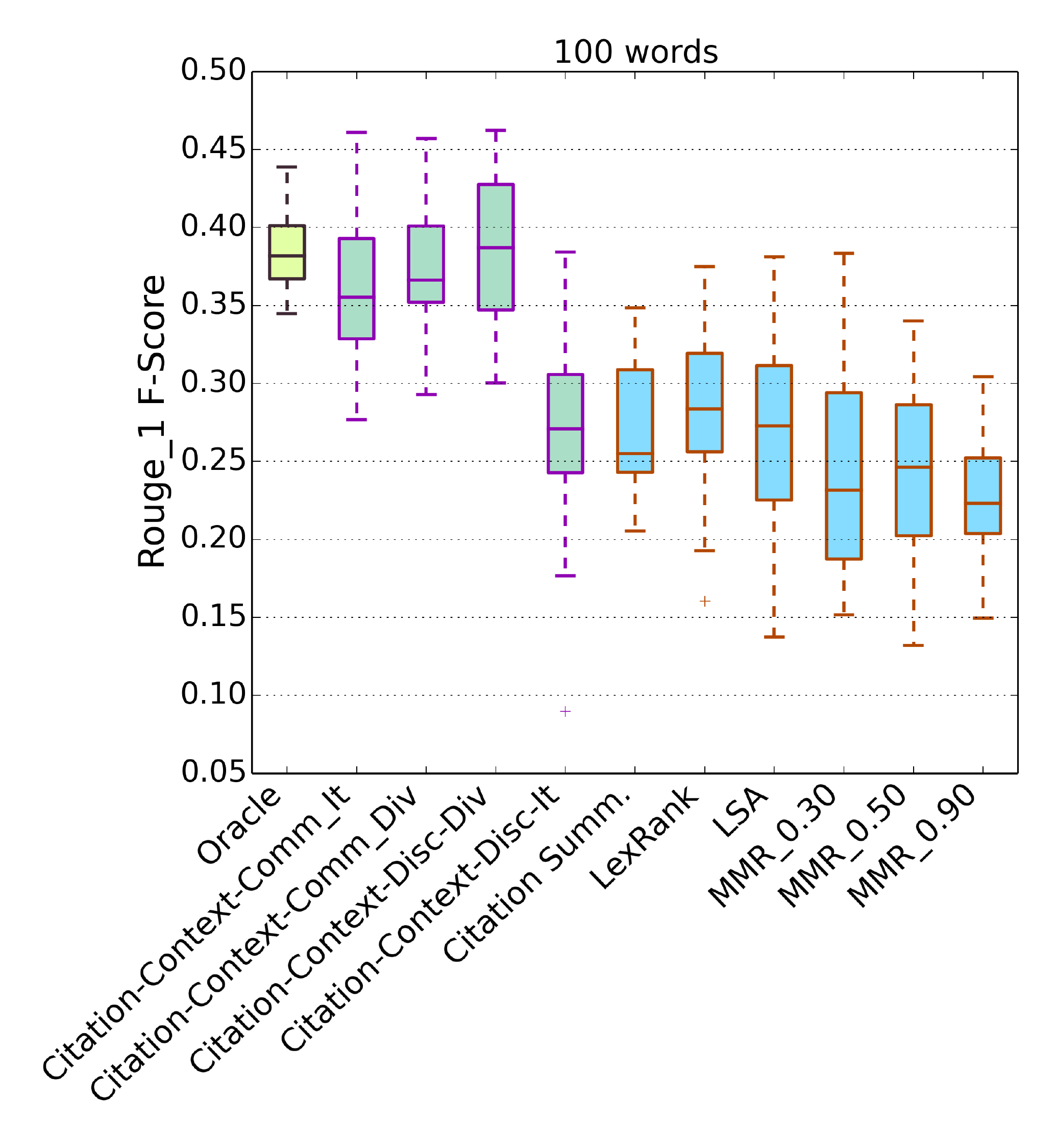}\hspace{4em}
\includegraphics[width=0.5\textwidth, height=0.295\textheight,keepaspectratio]{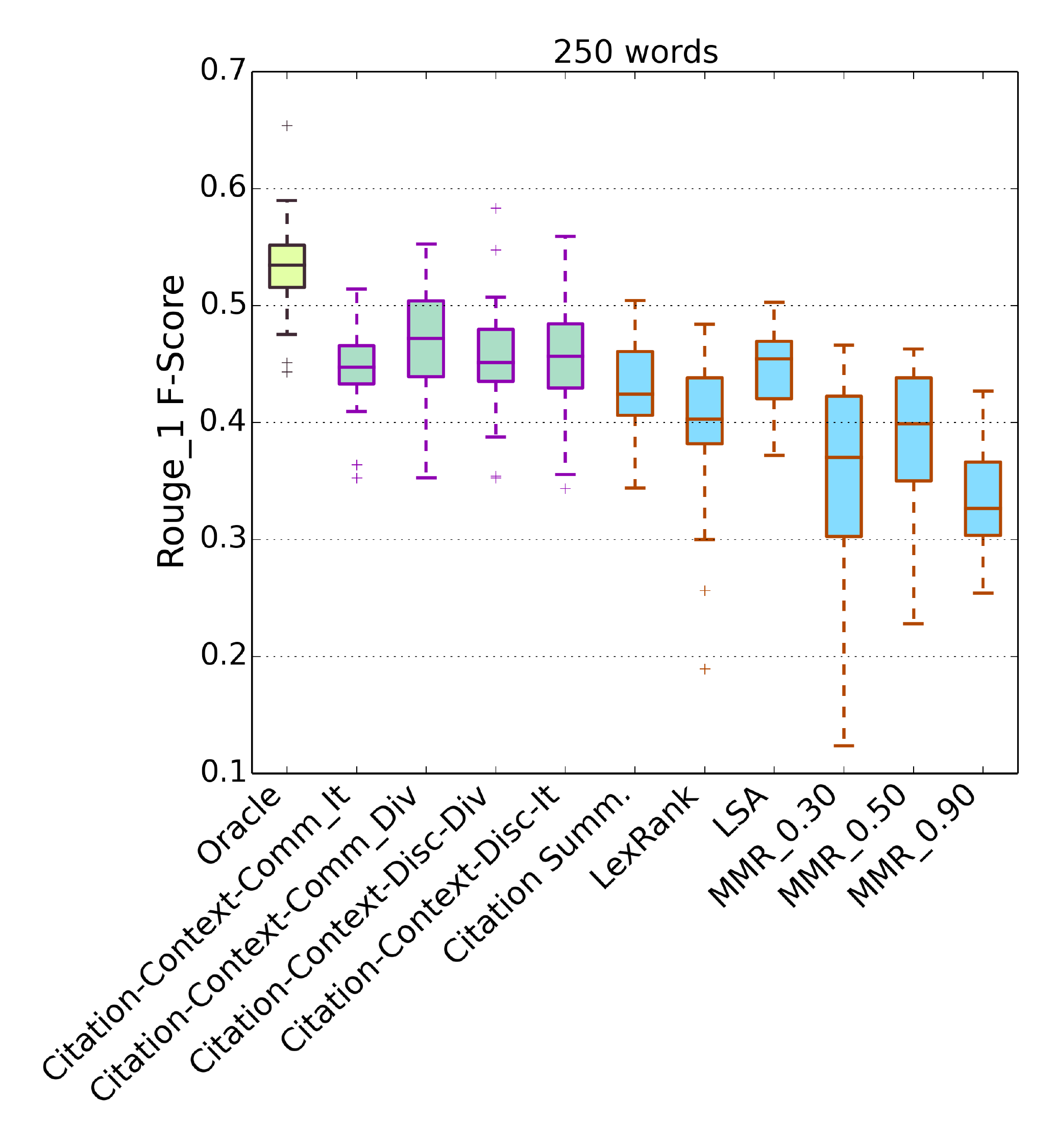}\hspace{0em}
\end{minipage} \hspace{0em}
\begin{minipage}[t]{\textwidth}
\hspace{1em}
\includegraphics[width=0.5\textwidth, height=0.295\textheight,keepaspectratio]{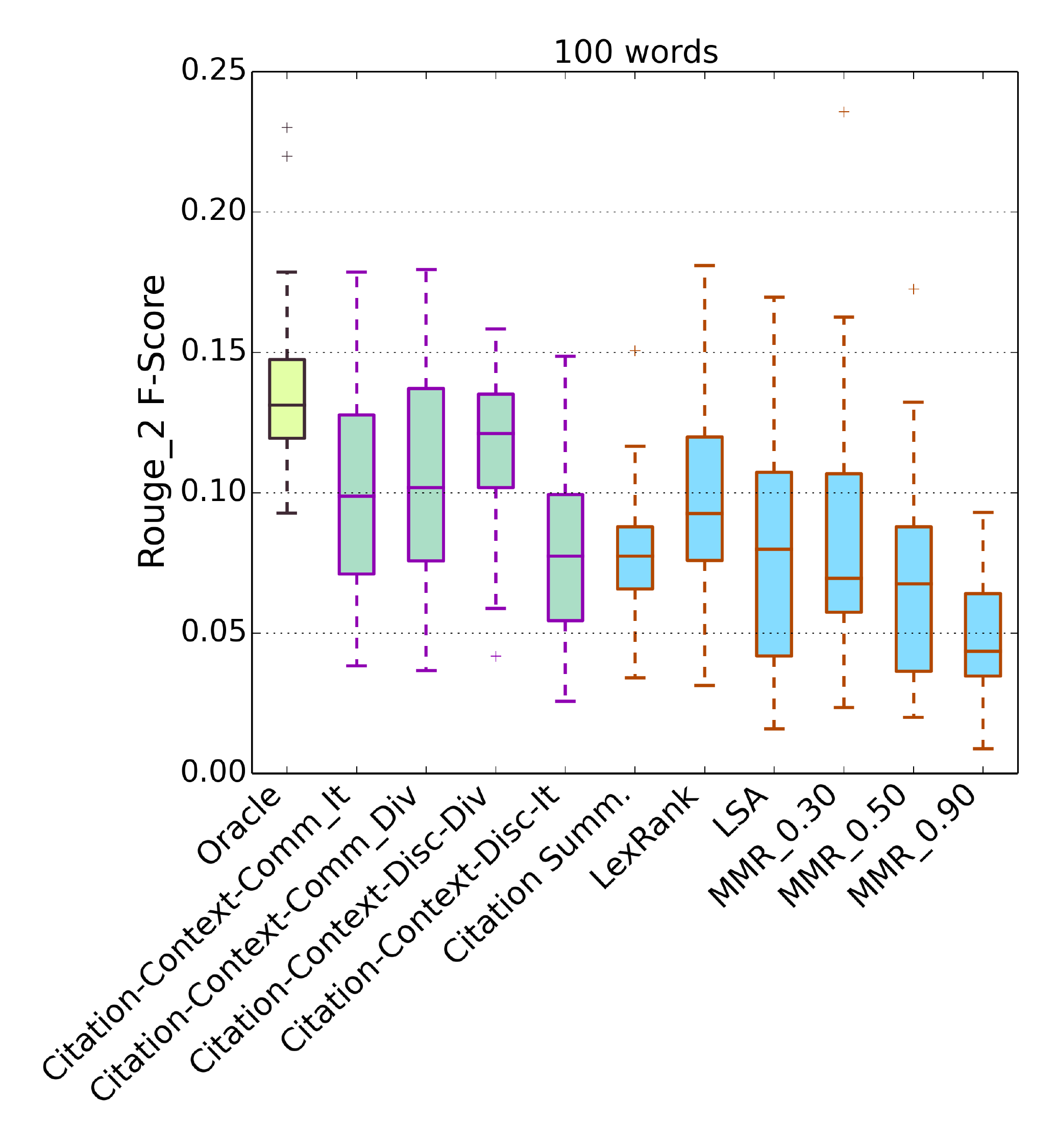}\hspace{4em}
\includegraphics[width=0.5\textwidth, height=0.295\textheight,keepaspectratio]{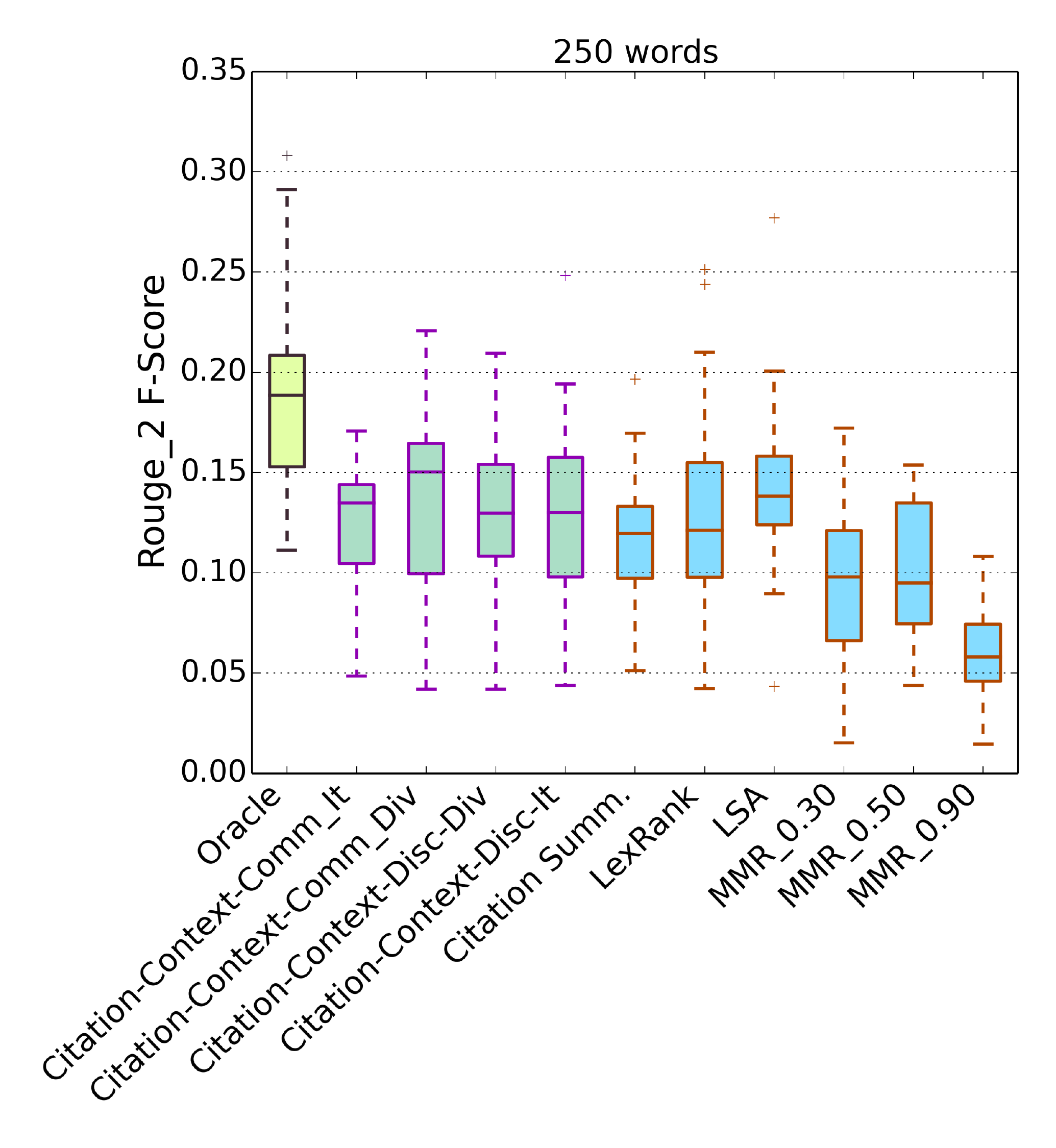}\hspace{0em}
\end{minipage} \hspace{0em}
\begin{minipage}[t]{\textwidth}
\hspace{1em}
\includegraphics[width=0.5\textwidth, height=0.295\textheight,keepaspectratio]{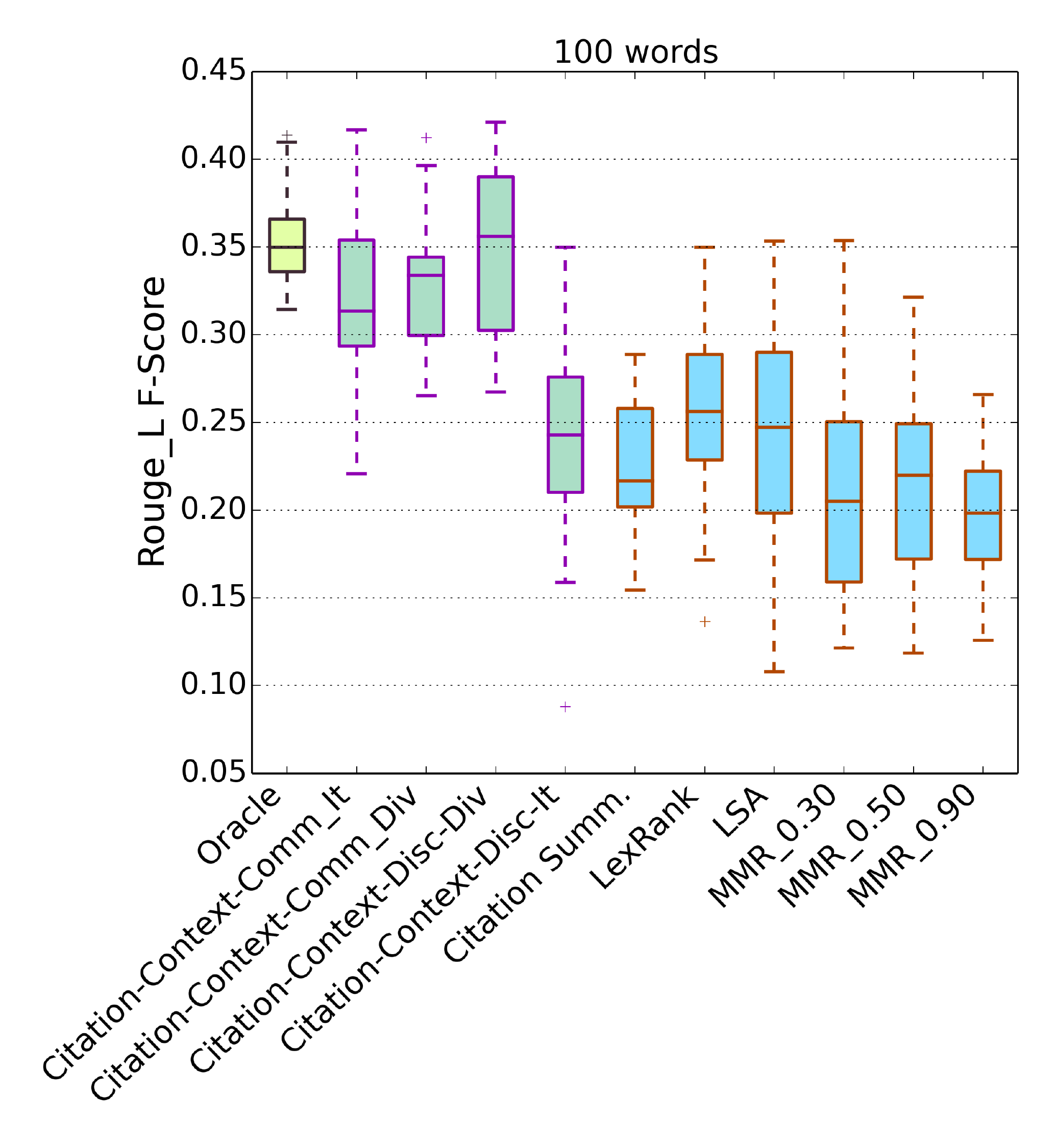}\hspace{4em}
\includegraphics[width=0.5\textwidth, height=0.295\textheight,keepaspectratio]{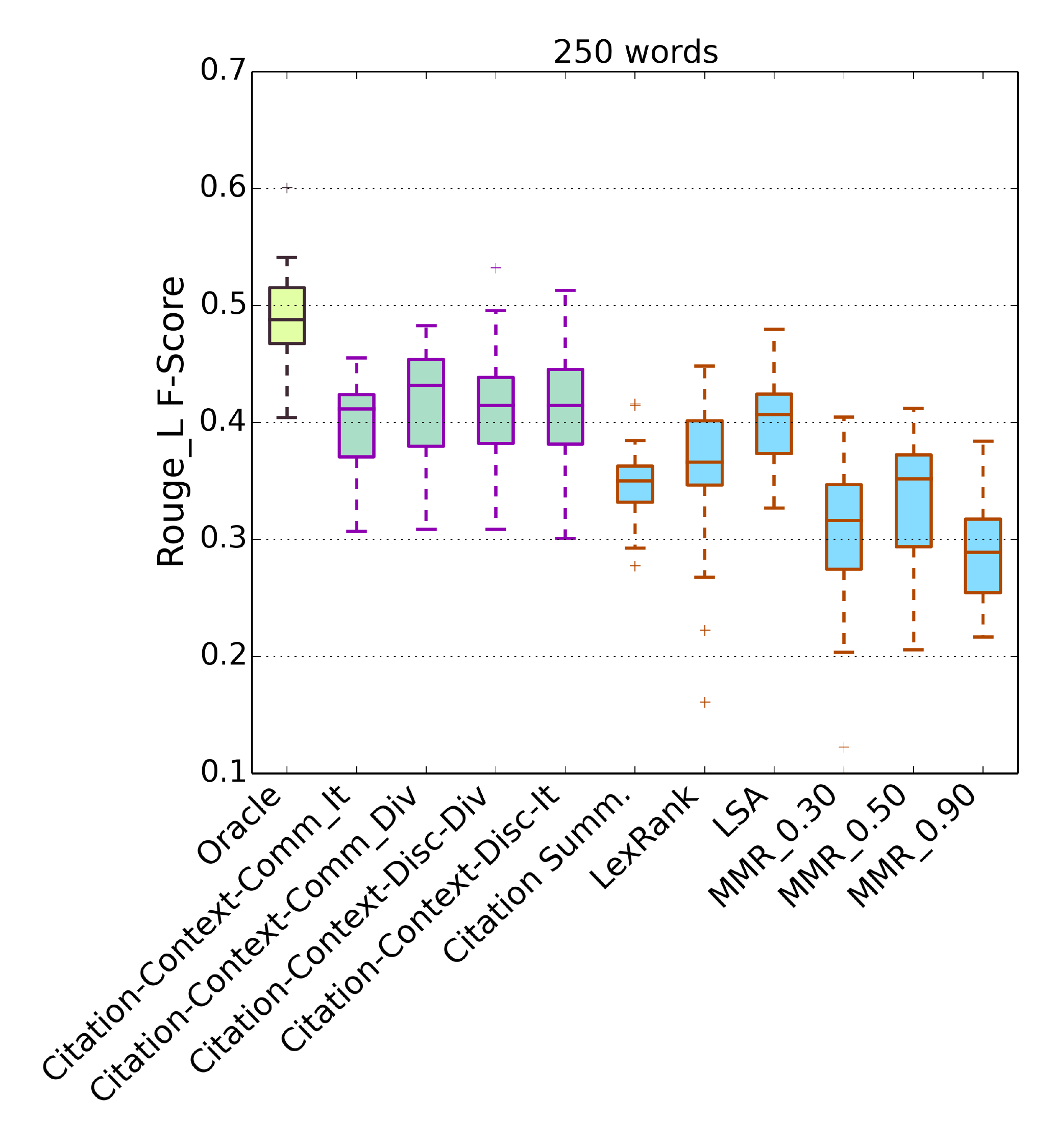}\hspace{0em}
\end{minipage}
\caption{\small   \textsc{Rouge-1}, \textsc{Rouge-2} and \textsc{Rouge-L} scores for different summarization approaches.
Chartreuse (yellowish green) box shows the oracle, green boxes show the proposed summarizers and blue boxes show the baselines; From left, Oracle; Citation-Context-Comm-It: Community detection on citation-context followed by iterative selection; Citation-Context-Community-Div: Community detection on citation-context followed by relevance and diversification in sentence selection; Citation-Context-Discourse-Div: Discourse model on citation-context followed by relevance and diversification; Citation-Context-Discourse-It: Discourse model on citation-context followed by iterative selection; Citation Summ.: Citation summary; MMR\_0.3: Maximal marginal relevance with $\lambda=0.3$. }
\label{fig:fig1}
\end{figure*}


We generated two sets of summaries using the methods and baselines described in previous sections. We consider short summaries of length 100 words and longer summaries of length 250 words (which corresponds to the length threshold in gold summaries). We also considered the oracle's performance by averaging over the \textsc{Rouge} scores of all human summaries calculated by considering one human summary against others in each topic. As far as 100 words summaries, since we did not have gold summaries of that length, we considered the first 100 words from each gold summary.
Figure \ref{fig:fig1} shows the box-and-whisker plots with \textsc{Rouge} scores. For each metric, the scores of each summarizer in comparison with the baselines for 100 word summaries and 250 words summaries are shown. The citation-context for all the methods were identified by the citation text vector method which uses the citation text except for numeric values, stop words and citation markers (first method in section \ref{subsec:method1}). In section \ref{sec:comp2}, we analyze the effect of various citation-context extraction methods that we discussed in \mbox{section \ref{sec:method}} on the final summary. The name of each of our methods is shortened by the following convention: [Summarization approach]\textunderscore[Sentence selection strategy]. Summarization approach is based on either community detection (Citation-Context-Comm) or discourse model of the article (Citation-Context-Disc) and sentence selection strategy can be iterative (It) or by relevance and diversification (Div).

We can clearly observe that our proposed methods achieve encouraging results in comparison with existing baselines. Specifically, for 100 words short summaries, the discourse based method (with 34.6\% mean \textsc{Rouge-L} improvement over the best baseline) and for 250 word summaries, the community based method (with 3.5\% mean \textsc{Rouge-L} improvement over the best baseline) are the best performing methods. We observe relative consistency between different rouge scores for each summarization approach. Grouping citation-context based on both the discourse structure and the communities show comparable results. The community detection approach is thus effectively able to identify diverse aspects of the article. The discourse model of the scientific article is also able to diversify selection of citation contexts for the final summary. These results confirm our hypotheses that using the citation context along with the discourse model of the scientific articles can help producing better summaries.

Comparison of performance of methods on individual topics showed that the citation-context methods consistently over perform all other methods in most of the topics (65\% of all topics).

While the discourse approach shows encouraging results, we attribute its limitation in achieving higher \textsc{Rouge} scores to the classification errors that we observed in intrinsic classification evaluation. In evaluating the performance of several classifiers, linear SVM achieved the highest performance with accuracy of 0.788 in comparison with human annotation performance. Many of the citations cannot exactly belong to only one of the discourse facets of the paper and thus some errors in classification are inevitable. This is also observable in disagreements between the annotators in labeling as reported by \cite{cohan2014}. This fact influences the diversification and finally the summarization quality.


\begin{figure*}
\begin{minipage}{\textwidth}
\includegraphics[width=0.33\textwidth, height=0.23\textheight]{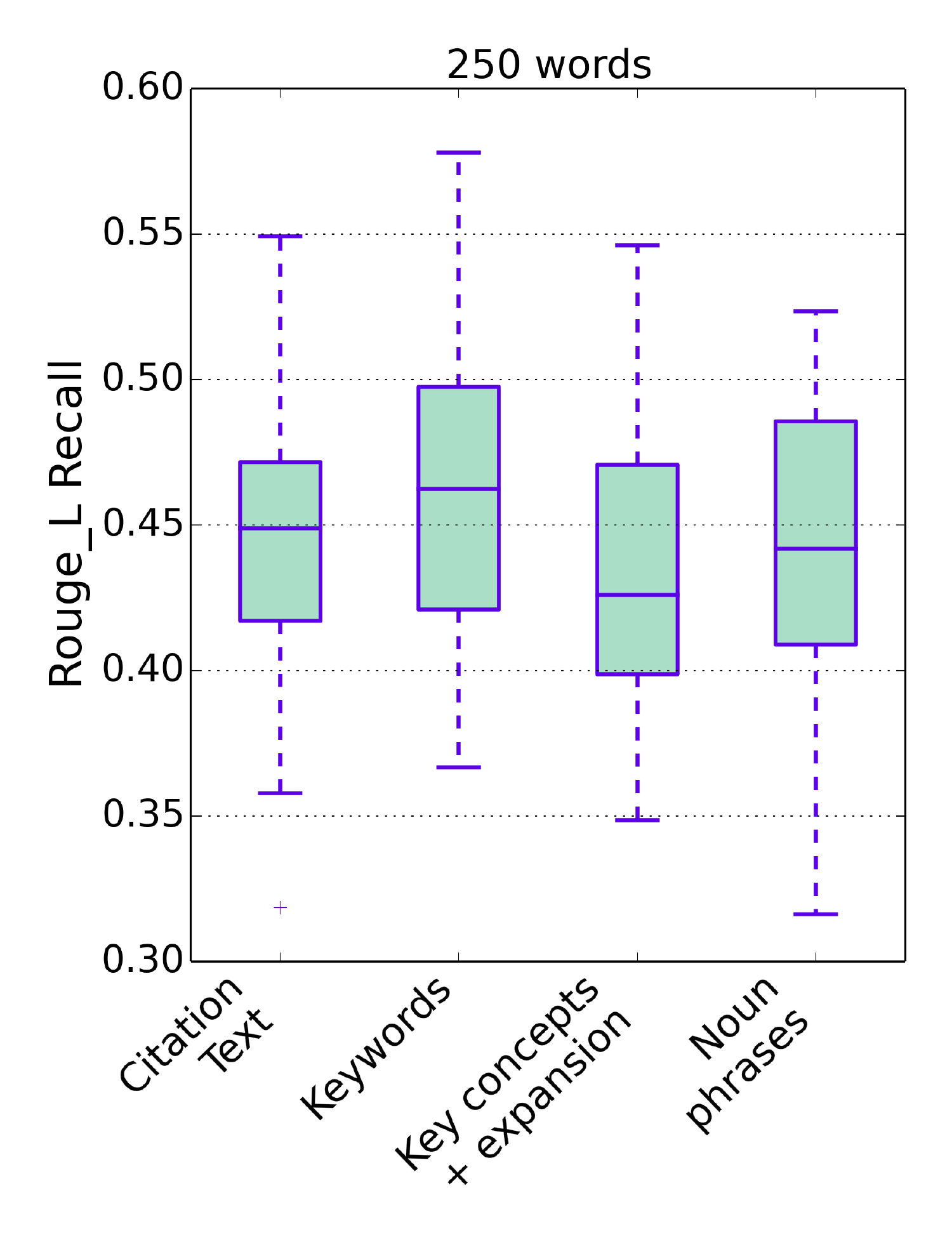}\hspace{-2pt}
\includegraphics[width=0.33\textwidth, height=0.23\textheight]{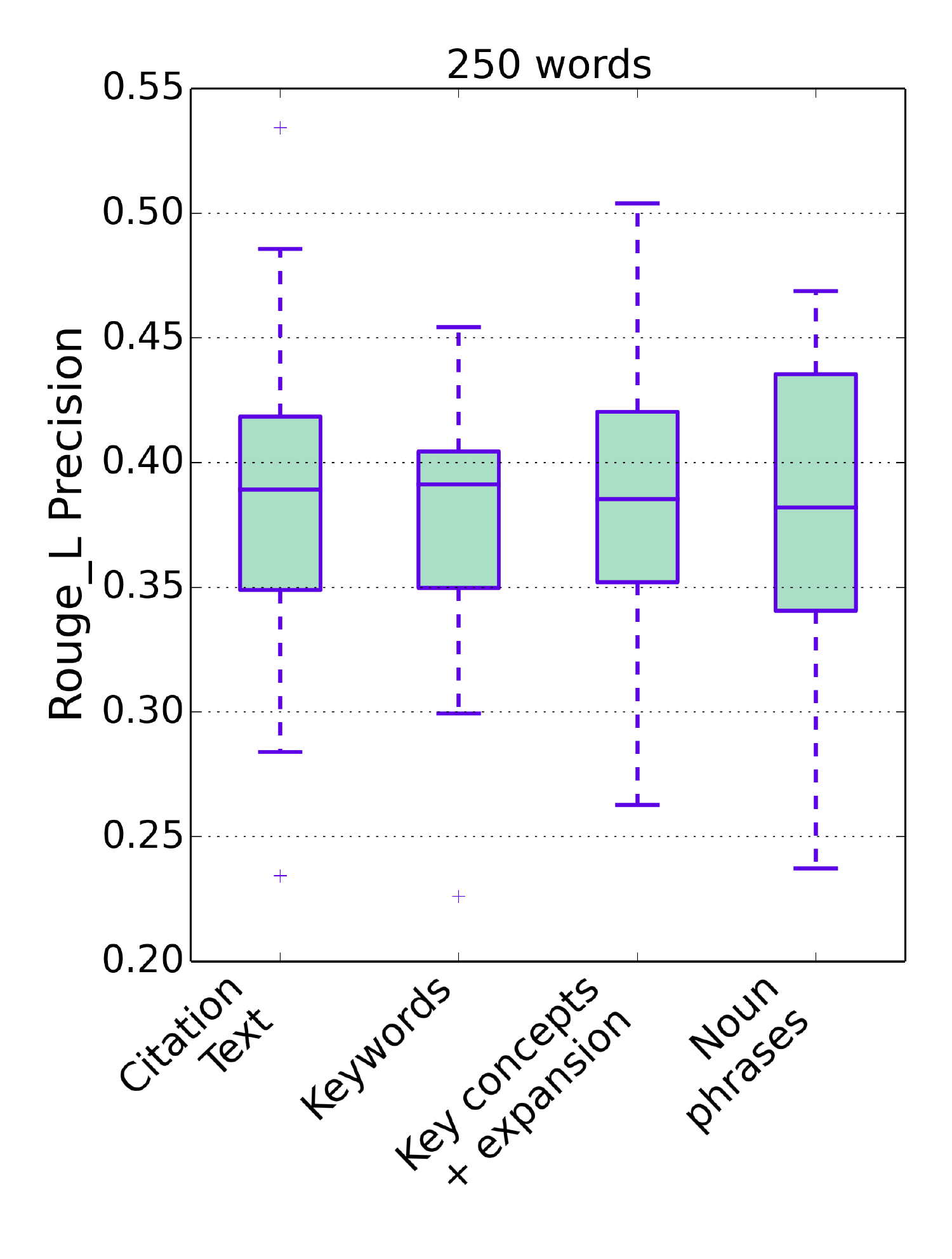}\hspace{-2pt}
\includegraphics[width=0.33\textwidth, height=0.23\textheight]{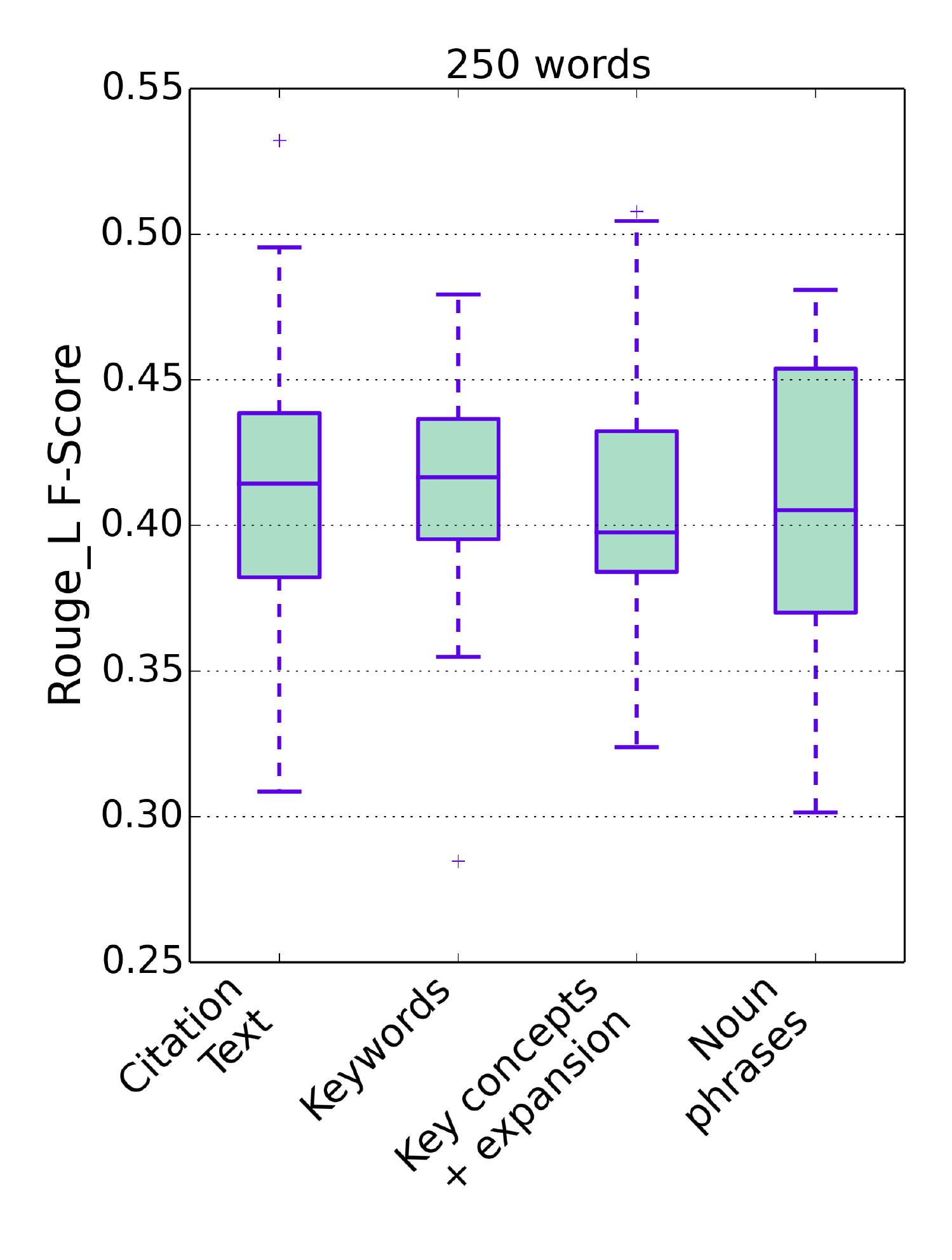}\hspace{0em}
\end{minipage}
\caption{\footnotesize Comparison of the effect of different citation-context extraction methods on the quality of the final summary.}
\label{fig:cit-extra}
\vspace{-2pt}
\end{figure*}

Among baseline summarization approaches, LexRank performs relatively well. Its performance is the best for short summaries among other baselines. This is expected since LexRank tries to find the most central sentences. When the length of the summary is short, the main idea in the summary is usually captured by finding the most representative sentence which LexRank can effectively achieve. However, the sentences that it chooses are usually about the same topic. Hence, the diversity in the gold summaries is not considered. This becomes more visible when we observe 250 word summaries. Our discourse based method can overcome this problem by including important contents for diverse discourse facets (34.6\% mean \textsc{Rouge-L} improvement for 100 words summaries and 13.9\% improvement for 250 word summaries). The community based approach achieves the same diversification effect in an unsupervised fashion by forming citation-context communities (27.16\% mean \textsc{Rouge-L} improvement for 100 words summaries and 14.9\% improvement for 250 word summaries).

The citation based summarization baseline has somewhat average performance among the baseline methods. This confirms that relying only on the citations can not be optimal for scientific summarization. While LSA approach performs relatively well, we observe lower scores for all variations of MMR approaches. We attribute the low performance of MMR to its sub optimal greedy selection of sentences from relatively long scientific articles.



By comparing the two sentence selection approaches (i.e., iterative and diversification-relevance), we observe that while for shorter length summaries the method based on diversification performs better, for the longer summaries results for the two methods are comparable. This is because when the length threshold is smaller, iterative approach may fail to select best representative sentences from all the groups. It essentially selects one sentence from each group until the length threshold is met, and consequently misses some aspects. Whereas, the diversification method selects sentences that maximize the gain in informativeness and at the same time contributes to the novelty of the summary. In longer summaries, due to larger threshold, iterative approach seems to be able to select the top sentences from each group, enabling it to reflect different aspect of the paper. Therefore, the iterative approach performs comparably well to the diversification approach. This outcome is expected because the number of groups are small. For discourse method, there are 5 different discourse facets and for community method, on average 5.2 communities are detected. Hence, iterative selection can select sentences from most of these groups within 250 words limit summaries.



\subsection{Analysis of strategies for citation-context extraction}
\label{sec:comp2}
Figure \ref{fig:cit-extra} shows \mbox{\textsc{Rouge-L}} results for 250 words summaries based on using different citation-context extraction approaches, described in section \ref{subsec:method1}.
Relatively comparable performance for all the approaches is achieved. Using the citation text for extracting the context is almost as effective as other methods. Keywords approach which uses the terms with high idf values for locating the context achieves slightly higher Rouge-L precision while it has the lowest recall. This is expected since keywords approach chooses only informative terms for extracting citation-contexts. This results in missing terms that may not be keywords by themselves but help providing meaning. Noun phrases has the highest mean F-score and thus suggests the fact that noun phrases are good indicators of important concepts in scientific text. We attribute the high recall of noun phrases to the fact that most important concepts are captured by only selecting noun phrases. Interestingly, introducing biomedical concepts and expanding the citation vector by related concepts does not improve the performance. This approach achieves a relatively higher recall but a lower mean precision. While capturing domain concepts along with noun phrases helps improving the performance, adding related concepts to the citation vector causes drift from the original context as expressed in the reference article. Therefore some decline in performance is incurred.


\section{Conclusion}
We proposed a pipeline approach for summarization of scientific articles which takes advantage of the article's inherent discourse model and citation-contexts extracted from the reference article\footnotemark[1]. Our approach focuses on the problem of lack of context in existing citation based summarization approaches. We effectively achieved improvement over several well known summarization approaches on the TAC2014 biomedical summarization dataset. That is, in all cases we improved over the baselines; in some cases we obtained greater than 30\% improvement for mean \textsc{Rouge} scores over the best performing baseline. While the dataset we use for evaluation of scientific articles is in biomedical domain, most of our approaches are general and therefore adaptable to other scientific domains.
\footnotetext[1]{Code can be found at: \url{https://github.com/acohan/scientific-summ}}

\section*{Acknowledgments}
The authors would like to thank the three anonymous reviewers for their valuable feedback and comments. This research was partially supported by National Science Foundation (NSF) under grant CNS-1204347.

\bibliographystyle{acl}
\bibliography{emnlp2015}

\end{document}